\documentclass[sigconf]{acmart}
\copyrightyear{2025}
\acmYear{2025}
\setcopyright{acmlicensed}
\acmConference[WWW '25] {Proceedings of the ACM Web Conference 2025}{April 28--May 2, 2025}{Sydney, NSW, Australia.}
\acmBooktitle{Proceedings of the ACM Web Conference 2025 (WWW '25), April 28--May 2, 2025, Sydney, NSW, Australia}
\acmISBN{979-8-4007-1274-6/25/04}
\acmDOI{10.1145/3696410.3714515}

\usepackage{amsmath}
\usepackage{algorithm, algorithmic}
\usepackage{natbib}
\usepackage{listings}

\begin{document}

\title{MoCFL: Mobile Cluster Federated Learning Framework for Highly Dynamic
Network}

%

\author{Kai Fang}
\affiliation{%
  \department{College of Mathematics and Computer Science}
  \institution{Zhejiang A\&F University}
  \city{Hangzhou}
  \state{Zhejiang}
  \country{China}
}
\email{Kaifang@ieee.org}

\author{Jiangtao Deng}
\affiliation{%
  \department{College of Mathematics and Computer Science}
  \institution{Zhejiang A\&F University}
  \city{Hangzhou}
  \state{Zhejiang}
  \country{China}
}
\email{dengjiangtao07@gmail.com}

\author{Chengzu Dong}
\affiliation{%
  \department{Division of Artificial Intelligence}
  \institution{Lingnan University}
  \city{Hong Kong}
  \country{China}
}
\email{chengzudong@ln.edu.hk}

\author{Usman Naseem}
\affiliation{%
  \department{School of Computing}
  \institution{Macquarie University}
  \city{Sydney}  
  \state{New South Wales}  
  \country{Australia}
}
\email{usman.naseem@mq.edu.au}

\author{Tongcun Liu}
\affiliation{%
  \department{College of Mathematics and Computer Science}
  \institution{Zhejiang A\&F University}
  \city{Hangzhou}
  \state{Zhejiang}
  \country{China}
}
\email{tongcun.liu@gmail.com}

\author{Hailin Feng}
\affiliation{%
  \department{College of Mathematics and Computer Science}
  \institution{Zhejiang A\&F University}
  \city{Hangzhou}
  \state{Zhejiang}
  \country{China}
}
\email{hlfeng@zafu.edu.cn}
\authornotemark[1]

\author{Wei Wang}
\affiliation{%
  \department{Artificial Intelligence Research Institute}
  \institution{Shenzhen MSU-BIT University}
  \city{Shenzhen}
  \state{Guangdong}
  \country{China}
}
\email{ehomewang@ieee.org}

\renewcommand{\shortauthors}{Kai Fang et al.}

\begin{abstract}
Frequent fluctuations of client nodes in highly dynamic mobile clusters can lead to significant changes in feature space distribution and data drift, posing substantial challenges to the robustness of existing federated learning (FL) strategies. To address these issues, we proposed a mobile cluster federated learning framework (MoCFL). MoCFL enhances feature aggregation by introducing an affinity matrix that quantifies the similarity between local feature extractors from different clients, addressing dynamic data distribution changes caused by frequent client churn and topology changes. Additionally, MoCFL integrates historical and current feature information when training the global classifier, effectively mitigating the catastrophic forgetting problem frequently encountered in mobile scenarios. This synergistic combination ensures that MoCFL maintains high performance and stability in dynamically changing mobile environments. Experimental results on the UNSW-NB15 dataset show that MoCFL excels in dynamic environments, demonstrating superior robustness and accuracy while maintaining reasonable training costs.
\end{abstract}



\maketitle

\section{Introduction}
With the rapid advancement of mobile cluster technology, application scenarios of highly dynamic networks, exemplified by unmanned aerial vehicle networks and vehicular networks, have been expanding rapidly, demonstrating great potential in fields such as emergency rescue, environmental monitoring, and communication infrastructure development \cite{al2024deep}. Mobile clusters not only enhance network coverage, flexibility, and reliability but also improve data collection and processing capabilities. However, the inherent dynamism of this technology presents several challenges. Frequent changes in terminal devices and network topology can complicate system management and potentially create security vulnerabilities \cite{peng2023tame}. Moreover, traditional centralized data processing methods are susceptible to privacy breaches and inefficiency due to the extensive collection and analysis of raw data.

FL, as a distributed machine learning paradigm, offers an effective solution to the challenges of data privacy and centralization. By training models locally and aggregating updates, this approach preserves data confidentiality while enabling collaborative model improvement. It integrates data from different devices without sharing raw data, reducing the risk of privacy leaks and minimizing data transmission needs. However, despite its strong performance in data privacy protection, existing FL still encounters limitations when applied to the complexities of highly dynamic network environments.

In highly dynamic mobile cluster networks, frequent topology changes and client churn intensify the non-independent and identically distributed (Non-IID) nature of data, significantly hindering the efficacy of existing FL approaches \cite{zhao2018federated}. The dynamic network environment leads to slower convergence in the federated aggregation process and diminished model accuracy.  Non-IID data distributions can introduce biases into model training, impacting the model's generalization capability. As illustrated in Figure 1, the frequent instability of terminals and network topology further exacerbates these challenges. \cite{chen2024survey}.

\begin{figure}[htbp]
\centering
\includegraphics[width=0.47\textwidth]{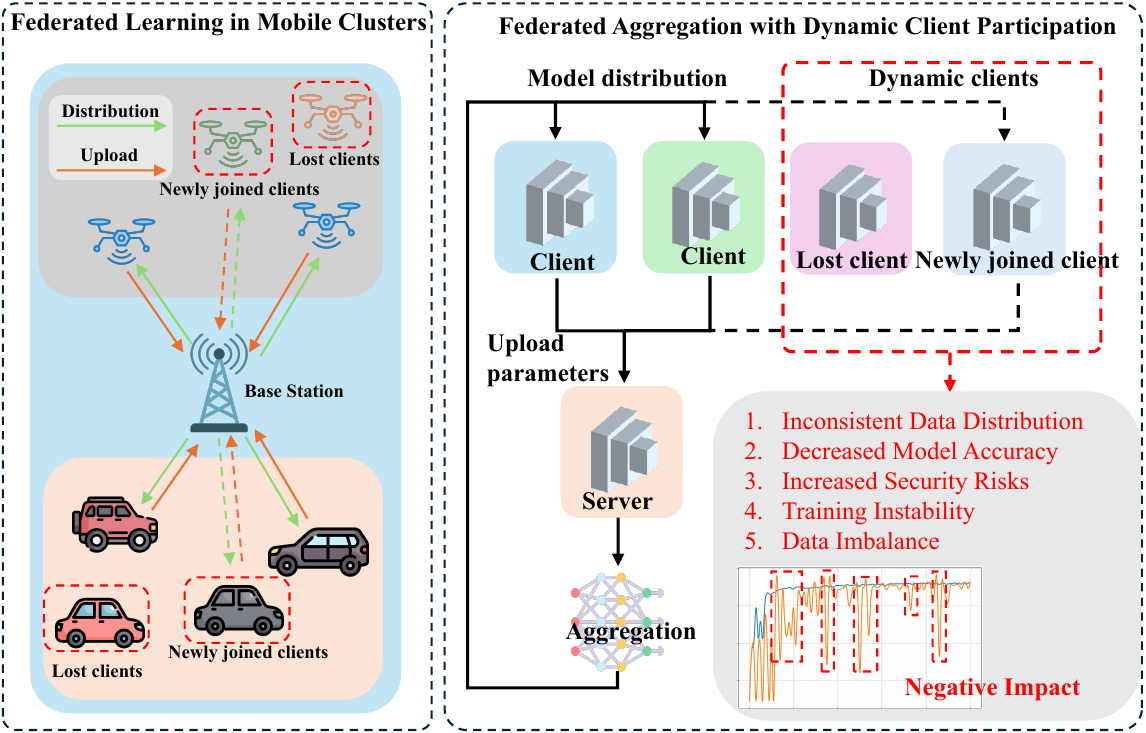}
\caption{
FL in Dynamic Mobile Clusters} \label{fig1}
\end{figure}

To tackle the challenges in highly dynamic mobile cluster networks, we propose MoCFL, an innovative FL framework designed for dynamic and heterogeneous environments. MoCFL uses an affinity matrix to assess the similarity between local feature extractors, selecting clients with similar data distributions for model updates. This enhances the model's adaptability to local data. The server aggregates feature representations to train a global classifier, ensuring generalized feature learning while preserving data privacy. MoCFL improves accuracy and reliability and provides a robust solution for security in dynamic mobile clusters. The main contributions of this paper are as follows:
\begin{itemize}
\item We introduce an affinity matrix to quantify client data similarities, enhancing knowledge sharing and improving feature extraction, leading to better test accuracy.

\item We use Maximum Mean Discrepancy (MMD) to measure divergence between feature representations, allowing the global classifier learn from both past and current data. This approach mitigates catastrophic forgetting and enhances generalization and robustness.

\item Experiments on the UNSW-NB15 dataset show MoCFL improves accuracy by 5-19\% over advanced static methods and maintains strong robustness in dynamic environments with reasonable computational overhead.
\end{itemize}

\section{Related Work}
\subsection{Federated Learning in Static Scenarios}

FL, as a crucial technology for data privacy protection and distributed computation, has seen widespread application and research in recent years. FedAvg \cite{mcmahan2017communication} is widely adopted for its efficient aggregation strategy, though it faces challenges with data heterogeneity. To mitigate these challenges, FedProx \cite{li2020federated} introduces a regularization term that improves model performance in heterogeneous environments. MOON \cite{li2021model} addresses non-IID data issues through a contrastive learning strategy during local training. FedAvgDBE \cite{zhang2024eliminating} promotes bidirectional knowledge transfer and reduces domain divergence between the server and clients. FedProto \cite{tan2022fedproto} employs a prototype-based approach to better align client models with a global prototype, enhancing the handling of data heterogeneity. FedAND \cite{kang2024fedand} utilizes the Alternating Direction Method of Multipliers (ADMM) to stabilize models against drift caused by server state changes. FedGH \cite{yi2023fedgh} addresses model heterogeneity by training a generalized global prediction head on the server, which is then used to replace local prediction heads on clients. FedBroadcast \cite{tian2022fedbroadcast} optimizes communication efficiency by broadcasting the global model and scheduling device updates, thereby improving convergence under extreme data distributions. FedCD \cite{liu2024fedcd} strategically combines centralized and decentralized architectures to reduce network bandwidth demands and accelerate training processes. VCDFL \cite{gou2024voting} enhances privacy protection by exchanging decision information rather than models, thereby reducing communication overhead. FedCORE \cite{li2024fedcore} proposes a privacy-aware collaborative training framework that addresses privacy concerns in cross-organizational scenarios by defining privacy leakage levels and offering robust protection strategies.

Although these FL strategies have effectively addressed challenges such as data and model heterogeneity, communication efficiency, and privacy in static environments, they often fall short in dynamic mobile settings. Frequent client changes, network instability, and device heterogeneity in mobile environments introduce new complexities that demand greater adaptability and resilience, which static FL strategies are not fully equipped to meet.

\subsection{Federated Learning in Dynamic Scenarios}
While static scenario FL methods have made significant progress in addressing data heterogeneity and privacy protection, applying them to mobile environments presents several challenges, such as network instability, device heterogeneity, and client mobility. To tackle these challenges, researchers have proposed various dynamic FL frameworks. DBFL \cite{khowaja2021toward} addresses connectivity and efficiency issues for remote devices by employing cluster protocols, making it compatible with mobile edge computing architectures.FedPE \cite{yi2024fedpe} optimizes local sub-networks adaptively based on system capacity each round, enhancing efficiency on resource-constrained mobile devices and reducing communication costs. AirCluster \cite{sami2023over} considers wireless communication characteristics and uses multi-model training and collaborative beamforming to alleviate communication bottlenecks, especially under severe data heterogeneity. Hou et al. \cite{hou2023efficient} applies FL to the metaverse by converting the problem into a Markov Decision Process with dynamic user selection and gradient quantization, optimizing performance and minimizing error bounds. MADCA-FL \cite{zhang2023vehicle} addresses vehicular networks by considering vehicle mobility and channel states, implementing FL while meeting delay and energy constraints. Additionally, ESVFL \cite{cai2024esvfl} achieves low-computation user-side encryption and verification through efficient privacy-preserving methods and cross-validation strategies, maintaining aggregation accuracy even with user dropouts.

Despite advancements in dynamic FL frameworks, challenges persist in managing significant fluctuations in data distribution caused by frequent client turnover in mobile environments. This instability in the feature space adversely affects model training, generalization, and overall performance. While existing research acknowledges client dynamics, systematic quantitative assessments are still lacking. Thus, ensuring model stability and accuracy in highly dynamic mobile environments remains a persistent challenge in FL. Further research is needed to develop more robust FL algorithms to address issues such as data drift, model inconsistency, and performance degradation resulting from frequent client changes.

\section{Methodology}
Traditional FL algorithms aggregate models from multiple FL clients via a central FL server to generate a global model. For instance, FedAvg \cite{mcmahan2017communication}, detailed in Algorithm 1, performs weighted aggregation of client models, assuming a high degree of homogeneity in local datasets. 
\begin{algorithm}[!h]
    \caption{Federated Averaging}
    \renewcommand{\algorithmicrequire}{\textbf{Input:}}

    \begin{algorithmic}[htbp]
        \REQUIRE Total number of clients $N$, number of selected clients $K$, fraction of clients that perform computation on each round $C$, number of communication rounds $T$, learning rate $\eta$, local batch size $b$, number of local epochs $E$. 
       
  Server executes: initialize  $\omega$.

 \FOR {each round $t=0$ to $T-1$} \STATE
    $K=\max (C \cdot N, 1)$
    
    $\mathcal{S}^t=(\text { random set of } K \text { clients })$

\STATE 
\STATE\textbf{Clients Side} (each client $k \in \mathcal{S}^t$):
\STATE Split local dataset $D_k$ into batches of size $b$: \STATE
$batches \leftarrow \text{split}(D_k, b)$
 \FOR {each local epoch  $i= 1$ to $E$} 
\FOR {batch $b \in \text {$batches$ }$} \STATE $\omega_k^t \leftarrow \omega_k^t-\eta \nabla \ell\left(\omega_k^t ; b\right)$
 \ENDFOR
 \ENDFOR
\RETURN $\omega_k^t$ to server
 \STATE

\STATE \textbf{Server Side}:
 \FOR {each client $k \in \mathcal{S}^t$ in parallel} \STATE 
 $\omega_k^{t+1} \leftarrow \omega_k^t$
 \ENDFOR
 \STATE$\omega^{t+1} \leftarrow \sum_{k=0}^{K-1} \frac{n_k}{n} \omega_k^{t+1}$
  \ENDFOR
    \end{algorithmic}
\end{algorithm}

\subsection{MoCFL Framework Overview}
\begin{figure*}[htbp]
\centering
\includegraphics[width=0.95\textwidth]{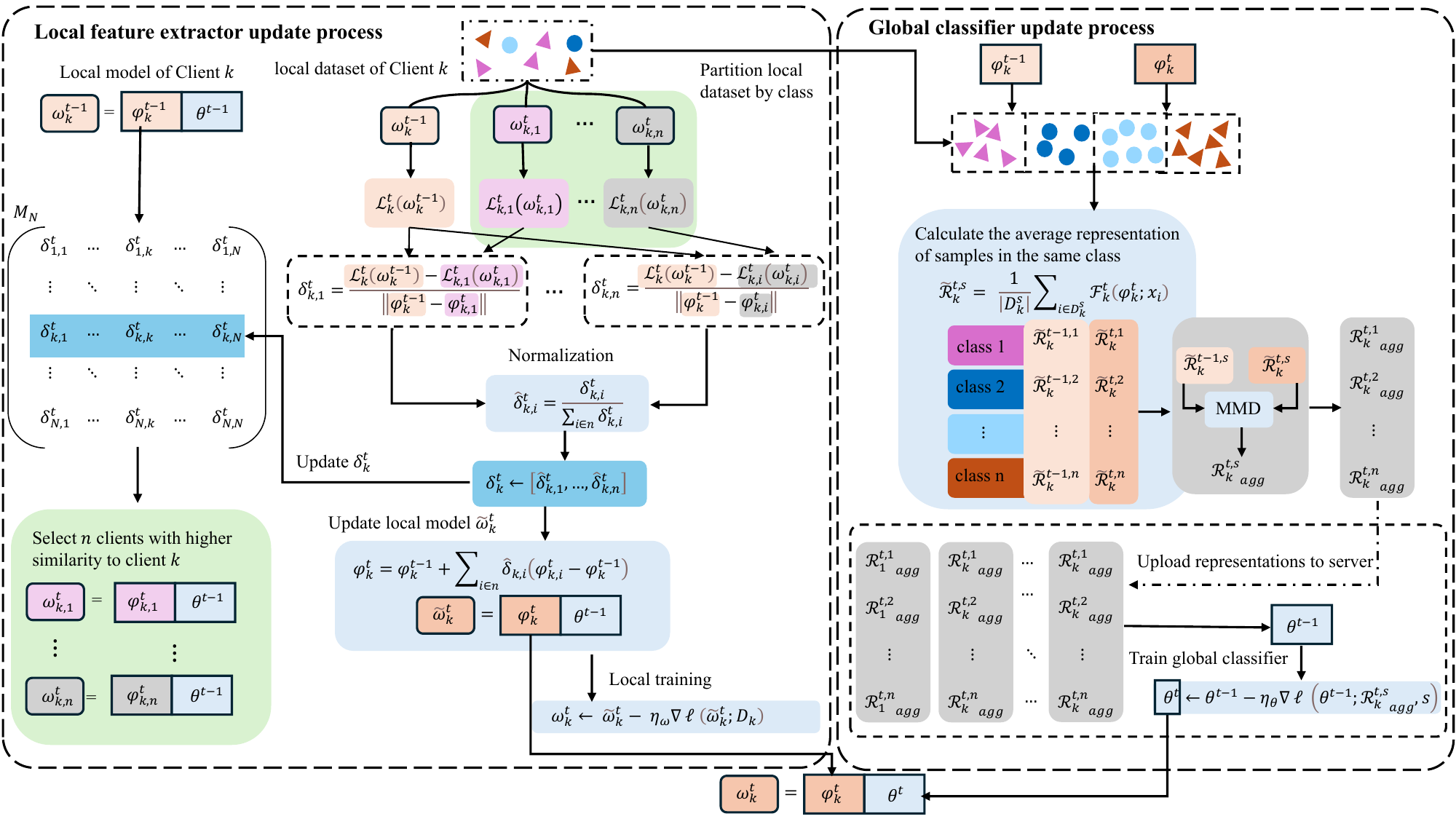}
\caption{
MoCFL System Framework} \label{fig2}
\end{figure*}

However, in mobile cluster environments, characterized by dynamic data distributions and high heterogeneity, the performance of FedAvg often falls short. To address this challenge, we propose MoCFL. In MoCFL, we first decompose the model into a feature extractor and a classifier. We then quantify the similarity between local clients by constructing an affinity matrix. Based on this similarity, we conduct knowledge transfer and aggregate local feature extractors to enhance feature extraction quality and improve recognition accuracy. Subsequently, we use the extracted feature representations to train the global classifier. During each iteration, the feature extractor extracts features from the client’s local dataset and collects the average feature representations from the previous and current rounds. MMD \cite{gretton2012kernel} is employed to measure the distributional differences between historical and current feature representations. These discrepancies are then used as weights in the aggregation process to produce updated feature representations, which are sent to the server. The server uses the combined features of historical and current feature representations to train the global classifier, which is then transmitted back to the clients. This approach ensures that the global classifier retains historical information, thus avoiding the loss of useful features and patterns, mitigating the forgetting problem in heterogeneous scenarios and enhancing the model's robustness. Moreover, training the global classifier with feature representations from all participating clients significantly improves the model's generalization capability.

Unlike traditional FL, MoCFL uses a personalized FL strategy, allowing each participant to customize their local model to better match their data and needs. By deploying targeted local models for each client, MoCFL allows for better adaptation to the individual local data distribution, as opposed to using a single global model. The global framework of MoCFL, shown in Figure 2, consists of two main components: local feature extractor updates and global classifier updates. The left side of Figure 2 illustrates the local feature extractor updates, where the extractor enhances aggregation by selecting similar feature extractors based on an affinity matrix, improving feature extraction for heterogeneous data. The right side depicts the global classifier updates, where the classifier is trained with both new and old feature information to address data heterogeneity and prevent forgetting, ensuring robust generalization. These components enhance feature extraction and global generalization, maintaining high robustness in dynamic mobile clusters. MoCFL seeks to minimize the total loss across all client models on their local datasets, formulated as:
\begin{equation}
    \min \sum_{k=0}^{N-1} \mathcal{L}_k\left(\omega_k\right),
\end{equation}
where $\mathcal{L}_k\left(\omega_k\right)$ represents the loss of the $k$-th client on its local dataset, and $\omega_k$ denotes the local model parameters of the $k$-th client. To achieve precise feature extraction and classification, we decouple the model $\omega$ into a feature extractor $\varphi$ and a classifier $\theta$, expressed as:
\begin{equation}
    \omega=\varphi \circ \theta,
\end{equation}
where $\varphi$ is the feature extractor, $\theta$ is the classifier, and $\circ$ denotes model concatenation. The system aggregates $\varphi$ to enhance feature extraction capabilities and trains the global classifier $\theta$ to improve the system's generalization performance. At the beginning of training, we randomly initialize the heterogeneous local models $\left[\omega_0^0, \ldots, \omega_{N-1}^0\right]$ and the global classifier $\theta^0$, while also setting $M_N$ as a diagonal matrix. In the $t$-th round of iteration, $K$ clients are randomly selected to participate in training, denoted as $\mathcal{S}^t$. Each selected client $k \in S^t$ selects the top $n$ most similar clients, denoted by $\mathcal{N}_k^t$, based on the weights from the affinity matrix $M_N$. Subsequently, client $k$ collect feature extractors $<\varphi_{k, 1}^t, \ldots, \varphi_{k, n}^t>$ from clients in $\mathcal{N}_k^t$. The server receives these feature extractor parameters and distributes them to the selected client $k$. The client combines its local classifier with the sampled feature extractors to form a complete model with different feature extractors:
\begin{equation}
    \begin{gathered}
\omega_{k, 1}^t=\varphi_{k, 1}^t \circ \theta^{t-1} \\,
\vdots \\
\omega_{k, n}^t=\varphi_{k, n}^t \circ \theta^{t-1}
\end{gathered}
\end{equation}

At this stage, each sampled client has $n+1$ models, including its local model $\omega_k^{t-1}=\varphi_k^{t-1} \circ \theta^{t-1}$. The client evaluates these models to update the weights in the affinity matrix $M_N$. The updated local feature extractor $\varphi_k^t$ is then computed as follows:
\begin{equation}
    \varphi_k^t=\varphi_k^{t-1}+\sum_{i \in \mathcal{N}_k^t} \hat{\delta}_{k, i}\left(\varphi_{k, i}^t-\varphi_k^{t-1}\right),
\end{equation}
where $\widehat{\delta}_{k, i}$ represents the normalized weight, which is used to compute the weighted sum of the differences between the feature extractors $\varphi_{k, i}^t$ and the previous round's feature extractor $\varphi_k^{t-1}$. The updated local feature extractor $\varphi_k^{t}$ is then combined with the previous round's global classifier $\theta_k^{t-1}$ to form a temporary local model $\widetilde{\omega}_k^t$. The local model $\omega_k^t$ is subsequently updated through gradient descent on the loss function as follows:
\begin{equation}
    \omega_k^t \leftarrow \widetilde{\omega}_k^t-\eta_\omega \nabla \ell\left(\widetilde{\omega}_k^t ; D_k\right),
\end{equation}
where $D_k$ denotes the local dataset held by client $k$, and $\eta_\omega$ is the local learning rate. After updating the local feature extractor, the aggregated new feature representations $\widetilde{\mathcal{R}}^{t-1, s}$ and $\widetilde{\mathcal{R}}^{t, s}$ are used to form the new aggregated feature representation $\widetilde{\mathcal{R}}^{t, s}{ }_{a g g}$. The server collects all new feature representations $\widetilde{\mathcal{R}}^{t, s}{ }_{a g g}$ to train the global classifier $\theta^t$ and distribute it to the clients participating in the next training round. 
\subsection{Update of Local Feature Extractors}
Clients with similar data distributions can optimize local model performance through knowledge sharing. To quantify the similarity between clients, we establish an $N$-dimensional client affinity matrix $M_N$. For the models $\omega_{k, 1}^t=\varphi_{k, 1}^t \circ \theta^{t-1}, \ldots, \omega_{k, n}^t=\varphi_{k, n}^t \circ \theta^{t-1}$ and $\omega_k^{t-1}=\varphi_k^{t-1} \circ \theta^{t-1}$ obtained in Eq. (4), we evaluate these $n+1$ models on the local model to update the weights in the affinity matrix $M_N$. First, we calculate the loss function value of the local model $\omega_k^{t-1}$, denoted as $\mathcal{L}_k^t\left(\omega_k^{t-1}\right)$. Next, we compute the loss function values $\mathcal{L}_{k, i}^t\left(\omega_{k, i}^t\right)$ for each $\omega_{k, i}^t$. The weight update formula is then applied as follows:
\begin{equation}
    \delta_{k, i}^t=\frac{\mathcal{L}_k^t\left(\omega_k^{t-1}\right)-\mathcal{L}_{k, i}^t\left(\omega_{k, i}^t\right)}{\left\|\varphi_k^{t-1}-\varphi_{k, i}^t\right\|},(i \in \mathcal{N}_k^t)
\end{equation}

The new weight is defined based on the difference between the loss function values $\mathcal{L}_k^t\left(\omega_k^{t-1}\right)$ and $\mathcal{L}_{k, n}^t\left(\omega_{k, i}^t\right)$, normalized by the $L_2$ norm of the difference between the feature extractor parameters $\varphi_k^{t-1}$ and $\varphi_{k, i}^t$. A larger difference indicates that the local model $\omega_k^{t-1}$ performs worse on the local validation set compared to the received model $\omega_{k, i}^t$, resulting in a higher weight for $\varphi_k^{t-1}$. 

To ensure numerical stability and avoid situations where the denominator approaches zero due to the cancellation of positive and negative terms, the result of Eq. (6) is processed with $\delta_{k, i}^t \leftarrow \max(0, \delta_{k, i}^t)$, followed by normalization of the positive values. This ensures that the computation remains well-defined and robust, particularly in cases where the differences in parameters are small.

When the difference between $\varphi_k^{t-1}$ and $\varphi_{k, i}^t$ is small, the shared knowledge can provide a greater benefit, leading to a higher weight for $\varphi_{k, i}^t$. Conversely, when the difference is large, the weight is relatively smaller, reflecting the limited potential improvement for the local model. Subsequently, normalization is applied to prevent instability caused by excessively large or small weights:
\begin{equation}
    \widehat{\delta}_{k, i}^t=\frac{\delta_{k, i}^t}{\sum_{i \in \mathcal{N}_k^t} \delta_{k, i}^t}
\end{equation}

The updated feature extractor $\varphi_k^t$ is obtained using the normalized weights as follows:
\begin{equation}
    \varphi_k^t=\varphi_k^{t-1}+\sum_{i \in \mathcal{N}_k^t} \hat{\delta}_{k, i}\left(\varphi_{k, i}^t-\varphi_k^{t-1}\right)
\end{equation}

After updating the weights $\left[\hat{\delta}_{k, 1}^t, \ldots, \hat{\delta}_{k, n}^t\right]$ for the $n$ clients similar to client $k$, the weight vector $\delta_k^t$ is updated as:
\begin{equation}
    \delta_k^t \leftarrow\left[\hat{\delta}_{k, 1}^t, \ldots, \hat{\delta}_{k, n}^t\right]
\end{equation}

The server then collects the updated weight vectors $<\delta_1^t, \ldots, \delta_k^t>$ from all participating clients and updates the affinity matrix $M_N$. This updated matrix facilitates the selection of similar clients for the subsequent iteration. By incorporating model updates from other clients and weighting the local feature extractors based on their relative contributions, the local feature extractors can learn relevant information more efficiently, thus accelerating model convergence and enhancing training performance.
\subsection{Update of Global Classifier}
To enhance the system's generalization and adaptability, we train the global classifier using feature representations. For client $k$, the updated feature extractors $\varphi_k^t$ and the previous feature extractor $\varphi_k^{t-1}$ are used to extract new and old feature representations, respectively. Feature representations $\mathcal{R}$ are numerical vectors that capture the essential characteristics of data segments. 

To align new and old features, we employ MMD to measure distributional differences. MMD is a statistical method that quantifies the discrepancy between two probability distributions, reflecting the differences between new and old feature representations. In the context of FL, where data distributions can be highly heterogeneous and complex, MMD effectively captures these distributional differences \cite{gretton2012kernel}. The MMD formula is: 
\begin{multline}
   MMD(P, Q)=\frac{1}{n_x^2} \sum_{p=1}^{n_x} \sum_{q=1}^{n_x} k\left(X_p, X_q\right)+\\ \frac{1}{n_y^2} \sum_{p=1}^{n_y} \sum_{q=1}^{n_y} k\left(Y_p, Y_q\right) + \frac{2}{n_x n_y} \sum_{p=1}^{n_x} \sum_{q=1}^{n_y} k\left(X_p, Y_q\right),
\end{multline}
where $P$ and $Q$ represent distributions $X$ and $Y$, $n_x$ and $n_y$ denote the number of samples in $X$ and $Y$, respectively, and $k(\cdot, \cdot)$ is the kernel function that measures the similarity between samples. This formula captures the differences between the internal self-similarity of distributions $X$ and $Y$ as well as their cross-similarity. 

Since feature representations are typically multi-dimensional, directly computing MMD significantly increases computational complexity and time cost. The time complexity for direct computation of MMD is  $O\left(n_x^2+n_y^2+n_x n_y\right)$. To reduce this complexity, we use an average feature representation approach. For the category $s$ in the local dataset $D_k$ of client $k$, the average feature representations are computed as follows:
\begin{equation}
    \widetilde{\mathcal{R}}_k^{t-1, s}=\frac{1}{\left|D_k^s\right|} \sum_{i \in D_k^s} \mathcal{F}_k^{t-1}\left(\varphi_k^{t-1} ; x_i\right),
\end{equation}

\begin{equation}
    \widetilde{\mathcal{R}}_k^{t, s}=\frac{1}{\left|D_k^s\right|} \sum_{i \in D_k^s} \mathcal{F}_k^t\left(\varphi_k^t ; x_i\right),
\end{equation}
where $\mathcal{F}_k^t\left(\varphi_k^t; x_i\right)$ represents the feature extraction using $\varphi_k^t$ on a subset of samples $x_i$ from the local dataset $D_k^s$. This process yields average feature representations $\widetilde{\mathcal{R}}_k^{t-1, s}$ and $\widetilde{\mathcal{R}}_k^{t, s}$ for categories $s$ in rounds $t-1$ and $t$, respectively. With average feature representations, Eq. (13) simplifies to:
\begin{equation}
    MMD\left(\widetilde{\mathcal{R}}_k^{t, s}, \widetilde{\mathcal{R}}_k^{t-1, s}\right)=2-k\left(\widetilde{\mathcal{R}}_k^{t, s}, \widetilde{\mathcal{R}}_k^{t-1, s}\right)
\end{equation}

This reduction in complexity simplifies the computation to\hfill\\$O\left(n_x+n_y\right)$, significantly lowering the cost compared to direct MMD computation.

The results derived from Eq. 13 are scaled to the range [0, 1], producing the normalized weights $W$, which reflect the relative importance of each local representation in capturing the underlying data distribution. These weights balance the contributions of historical feature representations $\widetilde{\mathcal{R}}_k^{t-1, s}$ and current feature representations $\widetilde{\mathcal{R}}_k^{t, s}$. We use weighted averaging to aggregate $\widetilde{\mathcal{R}}_k^{t-1, s}$ and $\widetilde{\mathcal{R}}_k^{t, s}$:
\begin{equation}
    \mathcal{R}_k^{t, s}{ }_{a g g}=W \cdot \widetilde{\mathcal{R}}_k^{t, s}+(1-W) \widetilde{\mathcal{R}}_k^{t-1, s},
\end{equation}
where $\mathcal{R}_k^{t, s}{ }_{a g g}$ represents the aggregated average representation for each local class in round $t$, used for training the global classifier. This process ensures that both current feature representations $\widetilde{\mathcal{R}}_k^{t, s}$ and historical representations $\widetilde{\mathcal{R}}_k^{t-1, s}$ are integrated to enhance the model's generalization ability and adaptability. By incorporating feature information from different time points, $\mathcal{R}_k^{t, s}{ }_{agg}$ avoids the inefficiency of directly transmitting entire models, particularly in resource-constrained environments. During aggregation, clients with feature distributions significantly deviating from historical distributions are assigned lower weights. This approach prevents the unified feature representation from being overly influenced by outliers or domain biases. Aggregating both old and new feature representations preserves historical information, mitigating the loss of useful features and patterns. This method helps maintain a memory of previously learned information, allowing better utilization of past experiences and knowledge.

On the server side, the aggregated feature representations $\mathcal{R}_k^{t, s}{ }_{agg}$ from all participating clients in the current round $t$ are used to train the global classifier. Parameter updates are performed by calculating the gradient of the loss function $\ell$ with respect to the global classifier $\theta^{t-1}$. The updated formula is:
\begin{equation}
    \theta^t \leftarrow \theta^{t-1}-\eta_\theta \nabla \ell\left(\theta^{t-1} ; \mathcal{R}_k^{t, s}{ }_{agg}, s\right),
\end{equation}
where $\eta_\theta$ is the learning rate for the global classifier, and\hfill\\ $\nabla \ell\left(\theta^{t-1} ; \mathcal{R}_k^{t, s}{ }_{agg}, s\right)$
 represents the gradient of the loss function $\ell$ with respect to the global classifier $\theta^{t-1}$, based on the aggregated feature representations $\mathcal{R}_k^{t, s}{ }_{agg}$ and category $s$. The updated global classifier $\theta^t$ is then distributed to the selected clients in the next round.

The global classifier integrates information from multiple models, enhancing prediction accuracy. By leveraging diverse perspectives, this approach mitigates individual model biases, improves robustness, and reduces overfitting, ultimately leading to better generalization.

\section{Experiments and Analysis}
To evaluate the effectiveness of the proposed MoCFL framework, comparative experiments were conducted on the UNSW dataset\cite{moustafa2015unsw} against seven advanced methods: FedAvg \cite{mcmahan2017communication}, FedProx \cite{li2020federated}, MOON \cite{li2021model}, FedAvgDBE \cite{zhang2024eliminating}, FedProto \cite{tan2022fedproto}, GPFL \cite{zhang2023gpfl}, and FedGH \cite{yi2023fedgh}. For all FL strategies, we employ the same 1D CNN model for intrusion detection. The model architecture comprises two convolutional layers, one global max pooling layer, one dropout layer, and one fully connected layer. Hyperparameters were standardized across all methods, with a local and global learning rate of 0.01, a batch size of 64, one local training epoch, and a total of 100 communication rounds. Experiments were executed on an RTX 4090 GPU using PyTorch 1.8.

\begin{figure*}[htbp]
\centering
\includegraphics[width=0.95\textwidth]{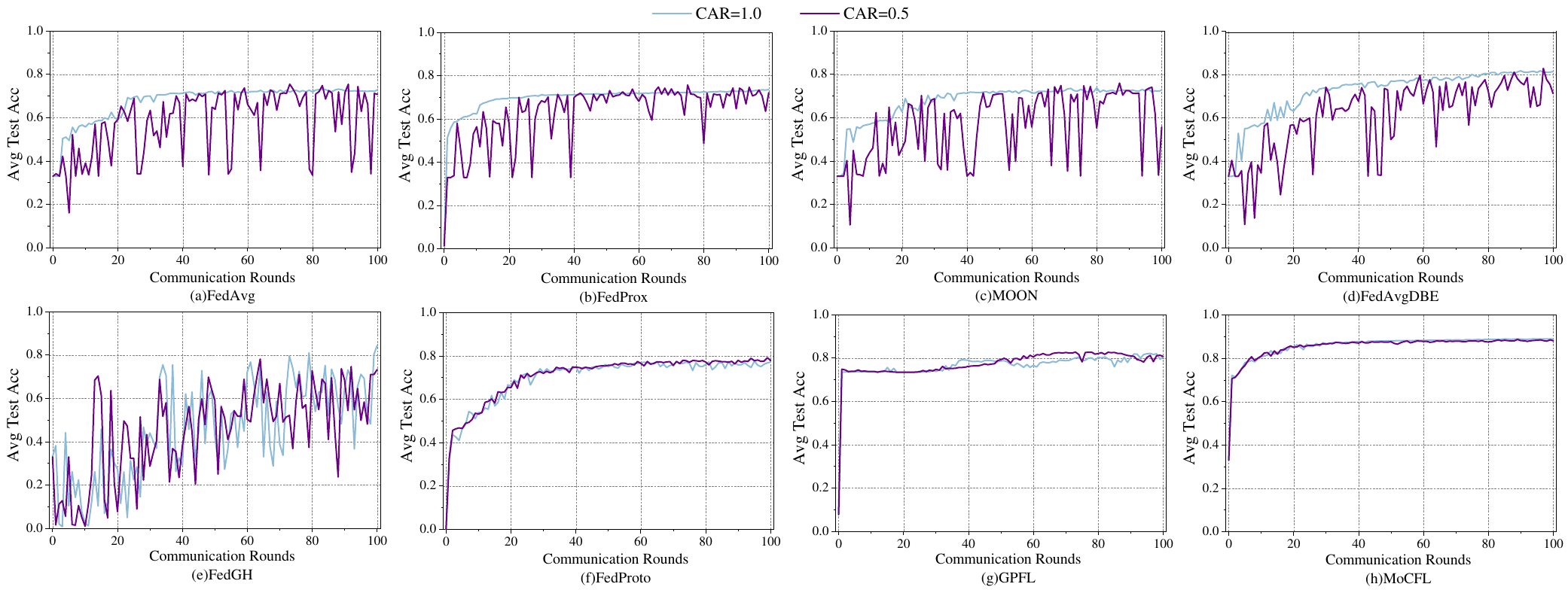}
\caption{
Performance of FL Strategies Under Varying Client Activity Rates in UNSW-NB15 Dataset (10 Clients)} \label{fig4}
\end{figure*}

\subsection{Comparison of Experimental Results in Static Scenarios}
In static scenarios with a fixed number of clients (NC), we compared the performance of various FL strategies. Table 1 presents the best accuracy rates achieved by different FL strategies under static conditions with varying client numbers, revealing significant performance disparities. Among all strategies, MoCFL achieved the highest accuracy, with rates of 0.8990 and 0.8645 for client numbers of 10 and 50, respectively, significantly outperforming other comparison algorithms. This superior performance underscores MoCFL's adaptability to highly heterogeneous static data. While FedGH, FedAvgDBE, and GPFL exhibited promising results, FedAvg, FedProx, and MOON struggled with highly heterogeneous data. The high accuracy of MoCFL is attributed to its personalized approach to local feature extraction and global classification. MoCFL enhances feature extraction by selecting clients with similar data distributions and model parameters for knowledge, facilitated by the affinity matrix. The global classifier effectively integrates both new and old feature information, enhancing the model's generalization across various static data scenarios.
\begin{table}[htbp]
\centering
\caption{Comparison of Best Accuracy Within 100 Communication Rounds}
\begin{tabular}{c|cc}
\toprule
Algorithm & NC=10  & NC=50  \\ \midrule
FedAvg                   & 0.7333 & 0.6781 \\
FedProx                  & 0.7359 & 0.6722 \\
MOON                     & 0.7387 & 0.6727 \\
FedAvgDBE                & 0.8175 & 0.7161 \\
FedProto                 & 0.7764 & 0.6963 \\
GPFL                     & 0.8216 & 0.7696 \\
FedGH                    & 0.8447 & 0.8238 \\
\textbf{MoCFL}                    & \textbf{0.8990} & \textbf{0.8645} \\ \bottomrule
\end{tabular}
\end{table}
These strategies collectively address data drift caused by data heterogeneity, leading to MoCFL's exceptional performance in static environments.

\subsection{Comparison of Experimental Results in Dynamic Scenarios}
To assess the impact of client dynamics on FL strategies, we introduced the client activity rate (CAR). During each aggregation round, only active clients update parameters, simulating client loss and rejoining. We evaluated each method's accuracy and robustness under different CARs. For a client population of 10, CAR values of 1.0 and 0.5 were tested. For a client population of 50, CAR values of 1.0, 0.9, 0.7, and 0.5 were evaluated. This comprehensive analysis provides insights into each FL strategy’s adaptability to dynamic client environments.

Figure 4 illustrates the impact of different CAR values on the accuracy of FL strategies with 10 clients. Only FedProto, GPFL, and MoCFL demonstrate robustness, while other methods experience significant performance drops at a CAR of 0.5. Moreover, when CAR is 1.0, these strategies have an average test accuracy below 80\% and exhibit frequent fluctuations. This lack of robustness is attributed to these methods’ inability to handle the high feature dimensions of the UNSW-NB15 dataset effectively. Among the robust methods, FedProto fails to achieve an average test accuracy of 80\%, while GPFL shows minor accuracy fluctuations after initially high precision, with minimal overall impact. In contrast, MoCFL maintains substantial robustness and high accuracy in this high-complexity scenario. At a CAR of 0.5, MoCFL retains an average test accuracy of 88\%, showing only a 1\% decline compared to the performance when CAR is 1.0. MoCFL's global classifier enhances its generalization ability, ensuring efficient intrusion detection even with high feature dimensions. Its dynamic adjustment mechanism further improves performance in complex scenarios. MoCFL utilizes the affinity matrix to evaluate client similarities, facilitating effective knowledge sharing and aggregation. This approach helps the system adapt to dynamic client changes. Additionally, the local feature extractors ensure the system maintains accuracy and stability even amid client churn. Consequently, MoCFL performs exceptionally well in high-complexity scenarios, maintaining high average recognition accuracy and stability.

\begin{figure*}[htbp]
\centering
\includegraphics[width=0.95\textwidth]{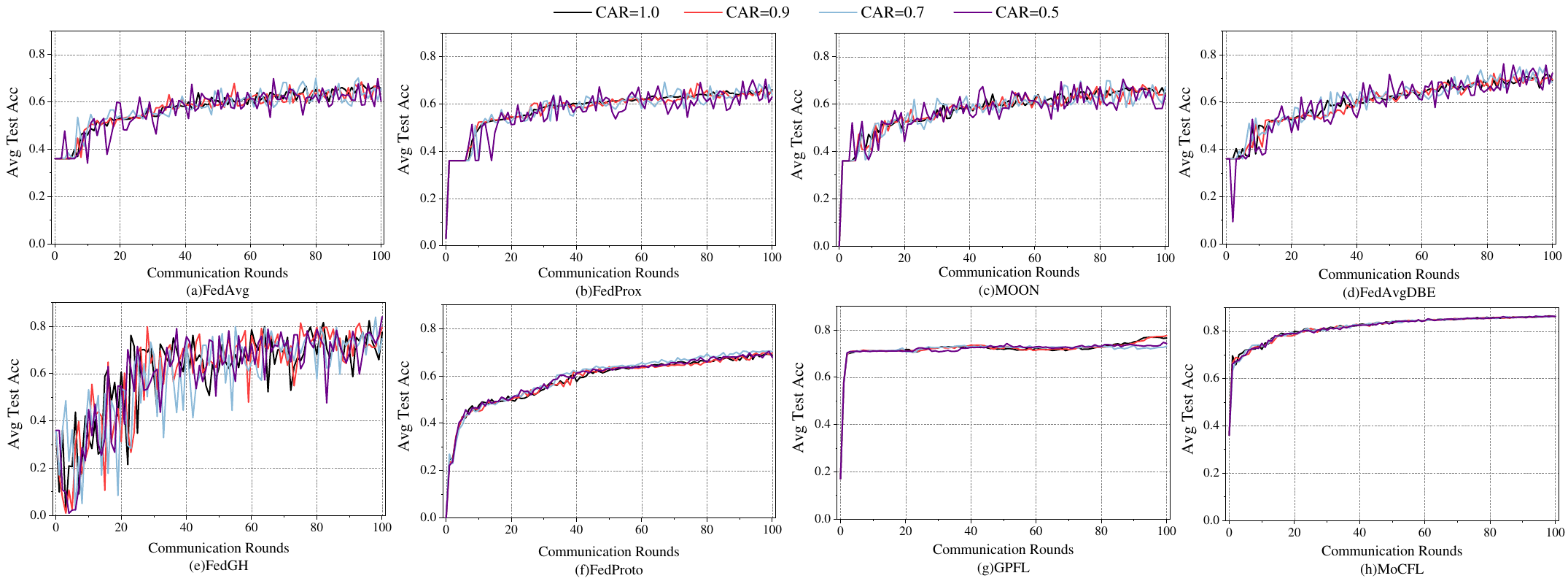}
\caption{
Performance of FL Strategies Under Varying Client Activity Rates in USNW-NB15 Dataset (50 Clients)} \label{fig5}
\end{figure*}

Figure 3 illustrates the impact of different CAR parameters on the performance of various FL strategies with 50 clients. FedAvg, FedProx, MOON, and FedAvgDBE fail to achieve an average test accuracy above 70\%. Although FedGH reaches a high accuracy, it struggles with significant fluctuations due to its reliance on the quality of feature representations, which its feature extractor fails to provide effectively. FedProto, despite having high robustness, achieves an average test accuracy of less than 70\%. GPFL achieves notable accuracy in early rounds but faces a performance bottleneck, with client activity influencing performance in the final rounds. Higher activity levels result in slightly better accuracy compared to lower activity levels. MoCFL stands out with the highest average test accuracy of 86\%, maintaining stable robustness throughout. Its unique optimization strategy and dynamic adjustment mechanism contribute to its exceptional performance, showing both the highest average test accuracy and robust stability.

Based on the results and analysis, the proposed MoCFL demonstrates the highest accuracy and robustness across various dynamic scenarios. MoCFL effectively addresses dynamic client changes and highly heterogeneous data through multiple optimization strategies. By evaluating client similarities using the affinity matrix, MoCFL enables knowledge sharing and strengthens aggregation, maintaining model stability and high accuracy when clients are lost or rejoin. Additionally, MoCFL aggregates feature representations before and after updates, merging historical and current information to address the forgetting problem caused by data heterogeneity. This approach ensures the stable extraction of high-quality features in complex scenarios. By leveraging high-quality feature representations to train the global classifier, MoCFL significantly enhances the model's overall generalization capability, maintaining high performance and stability despite variations in client activity levels.
\subsection{Time Overhead Analysis of Federated Learning Strategies}
In FL, training time overhead significantly affects the practical application and deployment of models. Table 2 presents the time overhead of each FL strategy over 100 communication rounds. 
\begin{table}[htbp]
\centering
\caption{Comparison of Time Overheads for 100 Communication Rounds}
\begin{tabular}{c|cc}
\toprule
Algorithm & NC=10  & NC=50  \\ \midrule
FedAvg                   & 186.16s & 812.41s \\
FedProx                  & 215.97s & 774.79s \\
MOON                     & 384.77s & 1569.32s \\
FedAvgDBE                & 242.82s & 984.09s \\
FedProto                 & 473.23s & 2025.88s \\
GPFL                     & 584.53s & 2419.57s \\
FedGH                    & 315.83s & 1315.73s \\
\textbf{MoCFL}                    & \textbf{327.61s} & \textbf{1247.55s} \\ \bottomrule
\end{tabular}
\end{table}

FedAvg and FedProx exhibit relatively short run times due to their update mechanisms, which primarily involve weighted averaging of client model parameters or the addition of regularization terms. These methods have lower computational complexity. However, these mechanisms may demonstrate reduced robustness and performance when dealing with highly heterogeneous data and dynamic clients. MOON and FedAvgDBE exhibit longer run times because of their additional feature representation and knowledge transfer mechanisms. MOON combines global and client model features for non-IID data, while FedAvgDBE improves knowledge transfer to reduce domain bias, increasing computational complexity. FedProto also has a longer run time, primarily due to its use of abstract class prototypes to reduce client-to-client differences, which involves extensive feature extraction and matching operations. Although this method effectively reduces information disparity between clients, it also increases the computational load, leading to higher time consumption. GPFL incurs increased computational overhead due to the integration of global and local features to enhance stability. While effective in handling client activity variations, this approach extends training time. FedGH has a moderate time overhead. Its characteristic reliance on a generalized global prediction head for feature representation leads to a more complex feature extraction process, resulting in intermediate time consumption. The proposed MoCFL achieves a favorable runtime balance among all strategies, with only a slight increase in runtime compared to FedAvg and FedProx. By evaluating client similarity using an affinity matrix, MoCFL facilitates efficient knowledge sharing and aggregation, reducing unnecessary computations. Additionally, MoCFL simplifies the computation of MMD during feature representation fusion, optimizing computational efficiency despite the added steps. In summary, MoCFL effectively controls runtime while maintaining high accuracy and robustness.

\section{Conclusion}
This study introduces MoCFL, a FL framework tailored for highly dynamic mobile networks. MoCFL uses hybrid optimization to enhance adaptability, dynamically assesses client similarity, and implements efficient knowledge sharing, maintaining accuracy despite frequent client changes. The affinity matrix ensures precise data integration, while the global classifier boosts generalization. Results on the UNSW-NB15 dataset show MoCFL's superior accuracy, robustness, and efficiency. In summary, MoCFL excels in handling heterogeneous data and dynamic networks, offering a robust solution for FL in mobile clusters and advancing intelligent data processing in complex environments.

\section{Acknowledgment}
This work was partly supported by the National Natural Science Foundation of China under Grant No. 62403433, the Natural Science Foundation of Zhejiang Province under Grant No. LQ23F020001, and the Quzhou City Science and Technology Project under Grant No. 2024K039.
\bibliographystyle{ACM-Reference-Format}
\balance
\bibliography{w4g25}

\appendix
\section{Dataset Description and Preprocessing}
The UNSW-NB15 dataset \cite{moustafa2015unsw} stands out as a comprehensive and realistic representation of complex network traffic scenarios, particularly those encountered in highly dynamic mobile cluster environments. This dataset captures a wide variety of network behaviors, including both normal traffic and malicious activities, making it highly suitable for evaluating intrusion detection systems and federated learning frameworks like MoCFL. It encompasses 10 distinct attack types, such as DoS, Exploits, and Worms, along with 49 features per record, which include flow-based, time-based, and content-based features. These features provide a rich and diverse foundation for assessing the adaptability and robustness of federated learning models in handling real-world network traffic. Figure 5 illustrates the distribution of data across the different attack types, highlighting the diversity and complexity of the dataset.

For data preparation, several preprocessing steps were undertaken to ensure the dataset was suitable for federated learning experiments. Categorical features, such as protocol type and service, were encoded into numerical values using one-hot encoding to facilitate model training. Numerical features were scaled using standardization to ensure consistent ranges across all features, which is critical for the convergence of machine learning algorithms. To simulate the non-IID (non-independent and identically distributed) nature of data in federated learning environments, the dataset was partitioned into subsets using the Dirichlet distribution with a concentration parameter 
$\alpha=0.3$. The Dirichlet distribution is a common choice for modeling non-IID data, as it allows for controlled variability in the distribution of classes across clients. By setting $\alpha=0.3$, we ensured a high degree of heterogeneity among client subsets, reflecting the diverse and imbalanced nature of real-world network traffic. This approach ensures that each client subset has a unique distribution of classes, with some clients potentially dominated by a single class while others exhibit a more balanced distribution. Each subset was further divided into training, validation, and test sets in a 4:1:1 ratio, ensuring a balanced evaluation of model performance across different phases of training and testing.

\begin{center}
\includegraphics[height=5.5cm,width=0.45\textwidth]{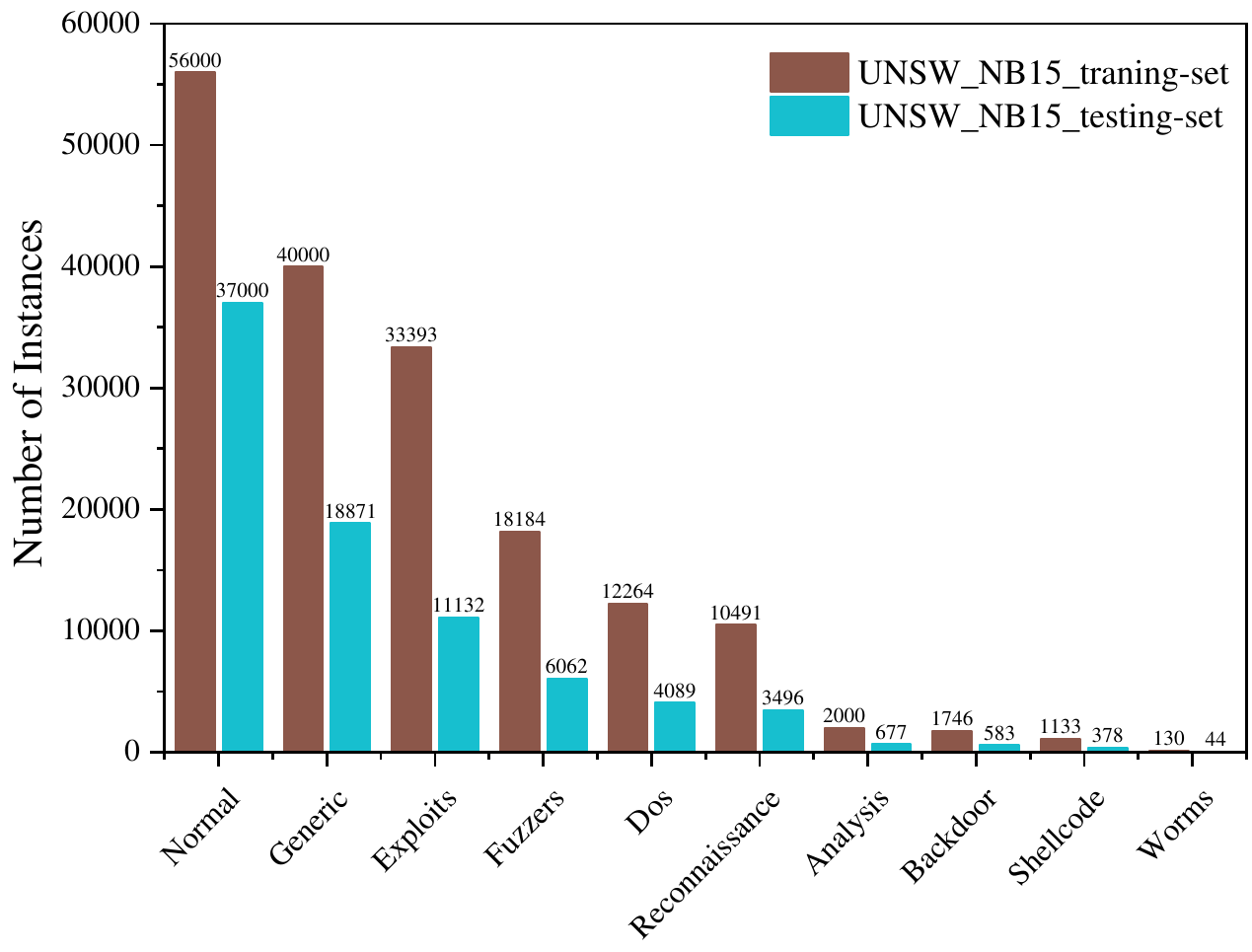}
\captionof{figure}{Distribution of Data Quantities for Different Attack Types in the UNSW-NB15 Dataset}
\label{fig2}
\end{center}

\begin{figure*}[htbp]
\centering
\includegraphics[height=4cm, width=0.95\textwidth]{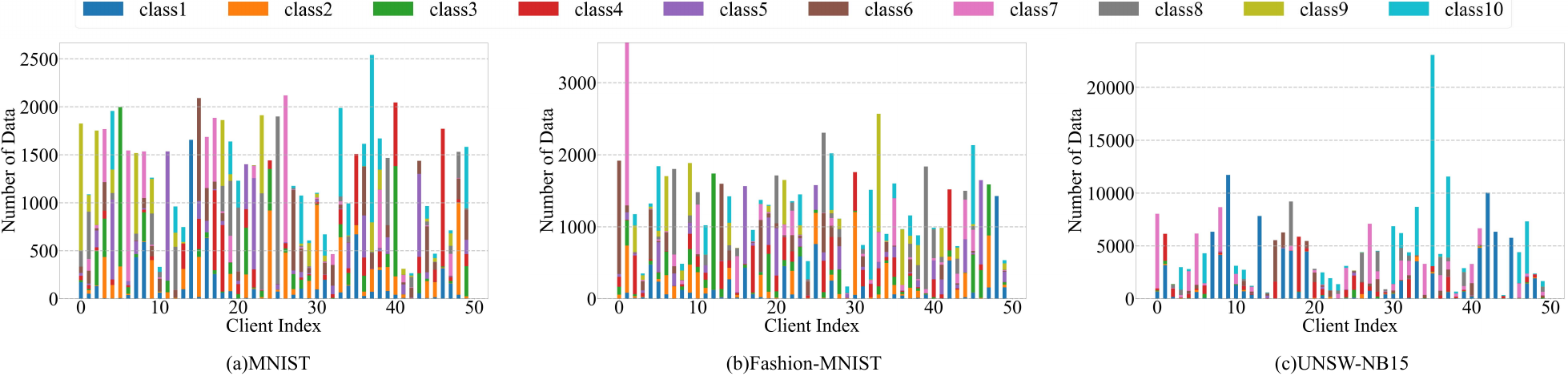}
\caption{Client Subsets under Dirichlet Distribution}
\label{fig6}
\end{figure*}

\begin{figure*}[htbp]
\centering
\includegraphics[height=4.2cm, width=0.95\textwidth]{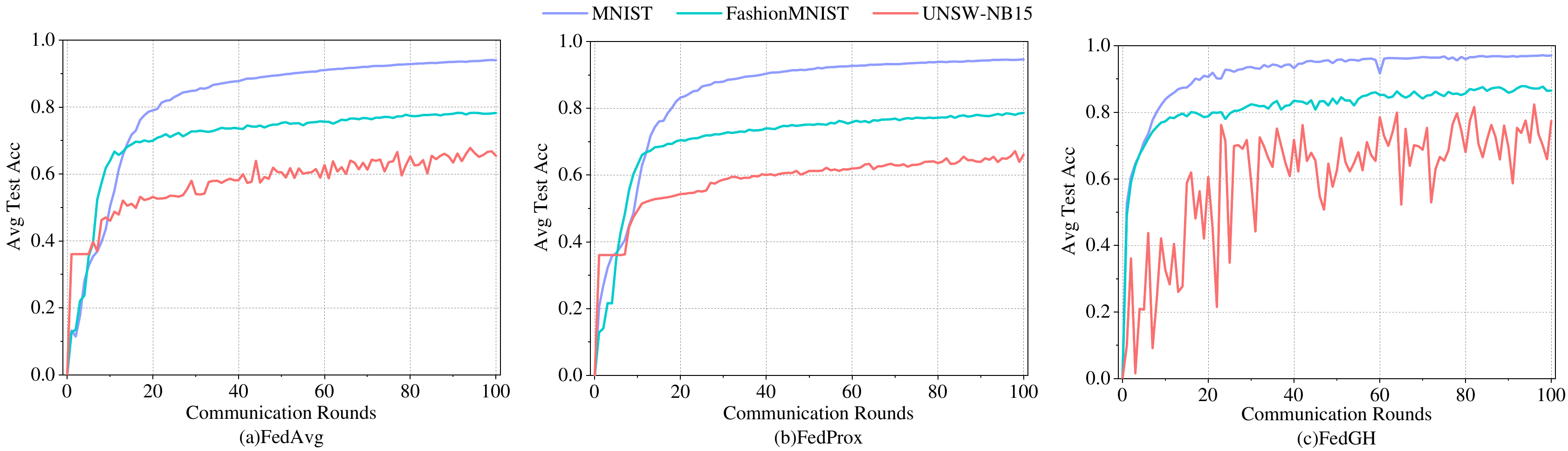}
\caption{Performance Comparison Across Different Datasets}
\label{fig7}
\end{figure*}
\section{Distribution of Dataset Classes and Impact of Imbalance on Federated Learning}

To gain deeper insights into how dataset characteristics influence federated learning performance, we conducted a comprehensive comparative analysis of three datasets: MNIST, Fashion-MNIST, and UNSW-NB15. MNIST and Fashion-MNIST are widely recognized benchmark datasets in federated learning research. In both datasets, each class represents approximately 10\% of the total samples, ensuring that no single class dominates the training process. This balance facilitates stable and consistent performance across federated learning algorithms.

In contrast, the UNSW-NB15 dataset presents a significantly more challenging scenario due to its extreme class imbalance. This dataset includes a diverse range of attack types, such as Fuzzers, Reconnaissance, and Shellcode, alongside normal traffic. However, the distribution of samples across these classes is highly uneven. For instance, the Worms category contains only 174 samples, constituting a mere 0.07\% of the total dataset. This imbalance becomes even more pronounced when the dataset is partitioned into client subsets for federated learning. Figures 6(a), (b), and (c) depict the distribution of client subsets for the three datasets under a Dirichlet distribution. While MNIST and Fashion-MNIST exhibit relatively uniform distributions across clients, the UNSW-NB15 dataset shows severe imbalances, with some client subsets containing critically scarce or excessively abundant samples of certain classes.

The impact of this imbalance on federated learning performance is evident in Figure 7, which compares the performance of three federated learning algorithms—FedAvg, FedProx, and FedGH—across the three datasets. While MNIST and Fashion-MNIST exhibit stable and consistent performance across all algorithms, with FedAvg achieving an accuracy of 97\% and 78\% respectively, the UNSW-NB15 dataset reveals significant challenges. In the UNSW-NB15 experiments, FedAvg exhibited noticeable fluctuations in accuracy, ranging from 58\% to 69\% across communication rounds, indicating instability due to the severe class imbalance. FedProx, which incorporates a proximal term to handle non-IID data, showed reduced fluctuations compared to FedAvg, with accuracy stabilizing around 64\% after 60 rounds, though it still struggled to achieve high performance. In contrast, FedGH experienced significant instability, with accuracy oscillating dramatically between 41\% and 82\% and failing to converge throughout the training process. These experimental results underscore the considerable negative impact of class imbalance on the performance of federated learning frameworks.

\clearpage

\end{document}